# Discover & eXplore Neural Network (DXNN) Platform, a Modular TWEANN.

*Gene I. Sher*

*Abstract*—In this paper I present a novel type of Topology and Weight Evolving Artificial Neural Network (TWEANN) system called Modular Discover & eXplore Neural Network (DXNN). Modular DXNN utilizes a hierarchical/modular topology which allows for highly scalable and dynamically granular systems to evolve. Among the novel features discussed in this paper is a simple and database friendly encoding for hierarchical/modular NNs, a new selection method aimed at producing highly compact and fit individuals within the population, a "Targeted Tunning" system aimed at alleviating the curse of dimensionality, and a two phase based neuroevolutionary approach which yields high population diversity and removes the need for speciation algorithms. I will discuss DXNN's mutation operators which are aimed at improving its efficiency, expandability, and capabilities through a built in feature selection method that allows for the evolved system to expand, discover, and explore new sensors and actuators. Finally I will compare DXNN platform to other state of the art TWEANNs on a control task to demonstrate its superior ability to produce highly compact solutions faster than its competitors.

## 1. INTRODUCTION

Neural Networks (NNs) are universal function approximators capable of modeling complex mappings between inputs and outputs. As the problem posed to the NN increases in complexity, the size of the NN must also increase, and as the size of the NN increases, the difficulty of training that NN increases with it. Even before the NNs reach moderate complexity and size, hand setting the weights, topology, and other parameters of the NN becomes impractical. The standard training algorithms like the error back-propagation[11] get stuck too easily in the local maximas of even just moderately difficult problems, like the single pole balancing task for example. An efficient solution for an automated method of setting both, the topology and the parameters in the Neural Network, is accomplished through the application of evolutionary algorithms (EA). By applying EA to evolve weights, general parameters, and topology of the NN, highly complex problems can be solved. The systems capable of topology and weight evolving through EA are referred to as Topology and Weight Evolving Artificial Neural Networks, or TWEANNs. There are many approaches and algorithms when it comes to TWEANNs, but even with these highly advanced systems the "curse of dimensionality" persists[12] and prevents some systems from solving the more complex problems. Thus the search for ever more sophisticated TWEANN approaches continues. One of such approaches is discussed in this paper.

I will present a new Neuroevolutionary system that through the use of hierarchy and modularity allows for an easier distribution of neural circuits among the computational hardware. The explicit modularity in the system provides a greater amount of flexibility by allowing for neural systems and non-neural systems to be easily integrated into one network. The proposed NN system referred to as: Discover & eXplore Neural Network (DXNN), presents a number of new features and improvements over the currently published monolithic TWEANNs. DXNN allows us to further accelerate the production of topological solutions to various problems, while at the same time consistently producing much more compact NNs when compared to even the most advanced TWEANN systems [3][4]. Furthermore, DXNN proposes a new approach for giving artificial life and robotic systems the ability to expand both sensory and actuator organs through built-in module based feature selection.

This paper shall be organized as follows: Section 2 will introduce the general functionality of TWEANN systems and list the modern state of the art TWEANNs with their short descriptions. Section 3 will provide a detailed description of the DXNN Platform and its encoding method. Section 4 will discuss DXNN platform's learning algorithm. Section 5 will discuss the results of the double pole balancing benchmark for DXNN and other TWEANNs. Section 6 will discuss the various possible advantages and disadvantages of explicitly modular NNs. Finally, section 7 will conclude with a summary and future plans for this platform.

Note: In this document, "DXNN Platform" refers to the entire software package which builds NNs, supervises and monitors the NN population, and applies the mutational operators to the NN. The hierarchical/modular NNs created by the DXNN Platform are referred to as DXNNs and NNs interchangeably.

## 2. MODERN TWEANN ALGORITHMS:

As the problem to be solved increases in complexity, a static NN becomes too inflexible to tackle it. If the NN is static, its size and topology must be guessed and hand designed beforehand by the researcher for every problem. Since this type of knowledge is very difficult to guess at from the problem itself and is impossible to guess when the problem reaches a non trivial level of complexity, a new automated method becomes a necessity. Furthermore, to mitigate the limitations of local search algorithms, an advanced system must be able to pull itself out of, or avoid altogether, the

[]Gene. I. Sher. Author is not associated nor was funded by any organization for the work and research presented in this paper.

local maximas that riddle the fitness landscapes of complex problems. TWEANN is a solution to such problems.

In TWEANN systems a seed population of individuals is created, each with some initial minimal or random topology of interconnected Neurons. Each NN within the population then attempts to solve the problem at hand, and based on its performance is assigned some fitness value. The individuals within the population are then compared with one another based on their fitness, and through some system specific algorithm the most fit of the individuals are allowed to produce offspring while the unfit individuals are discarded. The offspring of the successful NNs are either the mutated copies of the fit NN itself, or a combination of two or more NN through some crossover method. In this manner, through mutations or crossover, weights and topology of the NNs are changed, and through evolutionary pressure for higher fitness new and superior NNs are generated. As the complexity of the problem and the size of the NN grows, more and more weights, connections, and other parameters have to be set in just the right way to produce a solution. This problem is referred to as "the curse of dimensionality" and many state of the art TWEANN systems still become stuck in local maximas on the fitness landscape. Another problem is that some TWEANNs begin to add more and more neurons to the network while producing only very minimal fitness gain, this leads to increasingly bloated NN solutions. Ironically, this bloating results in more weights, parameters, and links that need to be set up concurrently to produce a solution, thus making it even harder, and usually impossible, for the NN to improve any further and solve the problem. The DXNN platform proposes algorithms to mitigate these difficulties.

### 2.1 Existing methods:

There are a number of existing algorithms that evolve NNs. Among such methods are the following state of the art systems: EPN[1], GNARL[2], NEAT[3], CoSyNe[4], HyperNEAT[5], EANT[6], and EANT2[7].

EPN: uses augmented back-propagation for weight optimization, and addition and removal of Neurons as topological mutation operators. EPN does not employ crossover.

GNARL: utilizes EA for both weight and topological optimization, utilizing a "temperature" parameter to determine the intensity of random mutations, a concept similar to the one used in simulated annealing [13]. Like EPN, GNARL also avoids utilizing cross-over.

NEAT: uses genetic algorithms to both mutate the weights and the topology. The weights are mutated through small perturbations, and the topological mutation operators are composed of: adding links to an existing Neuron, adding a new Neuron, crossover, and splicing. Splicing is described as the following process: two Neurons connected to each other are first disconnected and then reconnected through a newly created Neuron. NEAT also employs speciation, separating the NNs based on their topologies, trying to preserve diversity and allowing for the unfit individuals to survive for a few extra generations in hopes that they have the potential for improvement.

CoSyNe: uses a cooperative co-evolution approach, where various permutations of neurons belonging to different groups are tried in combination to determine which of them work best in most combinations. Based on such criteria the final NN is then composed by combining the best and most generally fit neurons.

HyperNEAT: is an extension of NEAT. In HyperNEAT, NEAT is used to drive a substrate of Neurons. The substrate of Neurons is composed of a neural grid. Each neuron on the grid has a coordinate distributed uniformly between -1 and 1 on all axis of the substrate. Instead of using the NN to solve the problem directly, HyperNEAT uses the NN to generate weights for the Neurons on the grid, also referred to as a substrate. The weight is generated for every Neuron by letting NN produce a weight based on the coordinate of a given Neuron and the coordinate of the Neuron linking to the given Neuron. When the substrate is a plane the problem becomes a 4 dimensional one with [X1,Y1,X2,Y2] being the input to the NN and [W] the output. Thus on a two dimensional substrate a weight W in a neuron A with coordinates [Xa,Ya] connected to from a neuron B with coordinates [Xb,Yb] is determined by feeding the NN the vector [Xa,Ya,Xb,Yb]. This type of indirect encoding has shown to produce interesting generalization capabilities. In this paper I shall refer to this type of encoding as Substrate Encoding (SE).

EANT2: separates the learning approach into two steps, exploitation through CMA-ES ("Covariance Matrix Adaptation Evolution Strategy")[8], and then exploration through standard topological mutations. It too does not utilize mating.

Neither CoSyNe nor CMA-ES evolve a topology, instead they optimize a single general topology. Thus these approaches can not be used for highly complex problems or open ended problems like Artificial Life. With these methods, the viable topological solution to the problems must be known beforehand.

### 3. DXNN Topology and Elements:

Modular DXNN platform was created with scalability and artificial life experiments in mind. It was created to be implemented by concurrent languages like MPI or OCaml, and to utilize the full power of distributed multi-core and multi-CPU hardware. The platform was made to produce modular NNs, a network of neural circuits with the ability to change the functionality of activation functions, linking methods, learning methods, and other features by simply updating the configuration parameters of the processing elements (Neurons, SubCores, and the Core). In DXNN each Node (Neuron, SubCore, or Core) is represented as an independent mini server/client with its own address and concurrency to all other nodes. This type of representation allows for the functionality of each element to be independent and on-line configurable. In the following sections I will first discuss the DXNN hierarchical architecture, and then follow by an elaboration on every element, its role, and its encoding.

### 3.1 THE ARCHITECTURE:

Modular DXNN is composed of 3 structural levels [Fig1]. At the lowest level are Neurons. When multiple Neurons are linked together to form some topological structure, this structure is referred to as a Neural Network or Neural Cluster, and each such NN is managed/supervised by a SubCore element. In DXNN a Neuron can utilize any type of activation function, such as a sigmoid, gaussian, sine, or any other. What input the NN gets and what its output is used for is determined by the SubCore element that governs and supervises that NN. SubCores represent the second level of hierarchy and themselves are linked together to form a network. Each SubCore supervises its own NN, passing it some input vector and gathering the resulting output vector. The SubCore Network itself is managed and supervised by a Core element. The Core is the top level structure and is the element which controls and deals with sensors and actuators, SubCores from/to signals, and the interface with OS. Core polls the sensors for data which come in vectors and passes those vectors to the appropriate input layer SubCores in the SubCore network. Each SubCore depending on its type, pre-process the vectors and passes them to its supervised NN. The SubCores then gather the processed signals from the Neurons in the output layers, post-process them to produce an output vector, and then pass those vectors onwards to other SubCores in the SubCore Network. The output layer SubCores pass their processed vectors to the Core. The Core gathers those vectors, packages each in a form appropriate for the actuator it is destined for, summons the actuator programs and passes each its own vector. This type of hierarchical approach allows not only explicit modularity by letting the SubCores form independent structures/modules, but also allows any program that can accept and output a vector to be regarded as a SubCore when embedded in a SubCore network. This type of topology also allows easy distribution of SubCores among separate CPU-cores, CPUs, or machines on a computer network in large experiments and projects. Explicit modularity minimizes the amount of overhead the system experiences as compared to a monolithic NN where concurrent Neurons are randomly distributed among the CPU-cores or machines. This is because each neuron by itself does little computation and outputs a single value, thus most of the processing becomes dedicated to the sending and receiving of tiny messages between CPUs/Machines. Using the SubCore modules allows us to distribute independent Neural Circuits among the machines or CPUs, and for single vectors belonging to the SubCores to be passed around rather than multiple single value vectors produced by the Neurons. To understand how DXNN Platform functions, I will first cover the functionality of each element (Neuron, SubCore, Core) and the flow of information within the system, and then follow up with the DXNN platform learning algorithm.

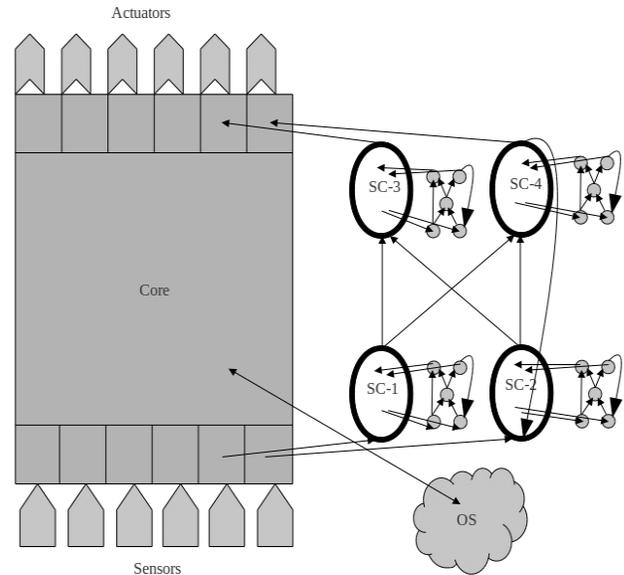

Fig. 1. The hierarchical structure of DXNN: Core, SubCores (SC-1...), and Neurons.

### 3.2 CORE:

Core is a Level 3 element. This element is the supervisor of the entire clustered Neural Network [Fig2], it is a program that is the interface between the SubCores and the OS/Environment/Sensors/Actuators. When live it is represented as a mini server/client with an Id/Address, a SensorList, an ActuatorList, a ParameterList that can further augment the Core's functionality, and a list of SubCore_Ids that the Core supervises. When stored in a database, it is represented as a tuple: {Id, SensorList, ActuatorList, ParameterList, SupervisedSubCoreIds, Generation, History}. SensorList is itself a list of tuples where each tuple is composed of a tag representing the name of a SensorProgram that the Core needs to run to get an input vector associated with a particular sense and an associated SubCore Id/Address which should receive the signals from that SensorProgram. SensorList can be represented as follows: [{SubCore_Id1, InfraredSensor_Id}...{SubCore_Idn, ChemicalSensor_Id}]. ActuatorList is a list of tuples where each tuple is composed of a tag representing a name of an ActuatorProgram and an associated SubCore Id/Address whose output vector is forwarded to the ActuatorProgram. ActuatorList can be represented as follows: [{SubCore_Id, LegsServoCluster_Id} … {SubCore_Idn, CameraTiltPanServoCluster_Id}]. When the SubCore sends a vector signal to the Core, the Core calls the associated ActuatorProgram and passes it the vector. The ActuatorProgram parses the vector and executes its function, whether it be moving a virtual agent, writing a value into a database, moving an actual robot by driving the servos, or modifying some part of the DXNN's topology.

<small>>This work has been submitted for possible publication.<</small> 4

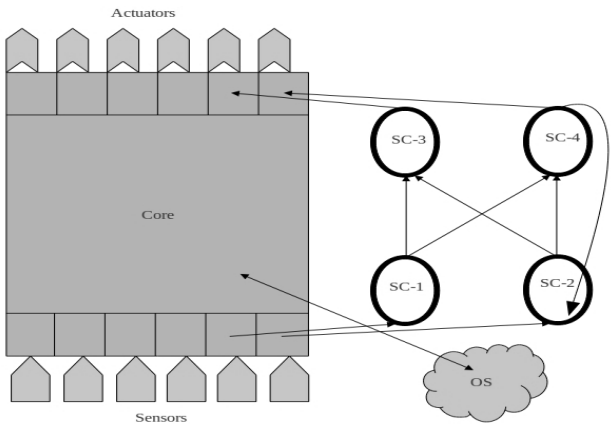

Fig. 2. Core, interfacing with sensors/actuators/subcores/OS.

For example, the Core might begin by going through the SensorList, calling the programs and passing the resulting vectors to the appropriate SubCores. A SensorProgram can be one that polls camera for data and then encodes an image as a vector of length n: [Val1,Val2,Val3...Valn] where Val is a scaled floating point. This vector is then passed to some designated SubCore for processing. At some later point SubCores in the output layer pass to the Core their output Vectors. Based on the Id/Address of the output layer SubCore, the Core chooses an ActuatorProgram associated with that SubCore, and then passes that actuator the Vector. The ActuatorProgram can for example control the servos to move a camera [Fig3] by sending it some signal which it derives by processing the vector. An example would be a vector: [Val1,Val2], which can represent pan and tilt signals respectively. The ActuatorProgram might further scale the Values within the vector to the appropriate ranges for the hardware it controls. The "Generation" variable is an integer that increments every time the DXNN goes through a topological mutation phase. "History" is a list composed of all the mutations applied to the DXNN listed in the order they were applied. The History list is composed of the following tuples: {MutationOperator, ElementAppliedTo, Info}. Where the MutationOperator is a tag/name of the mutation operator, ElementAppliedTo is an Id, and Info is extra information, if any, and depends on the type of MutationOperator.

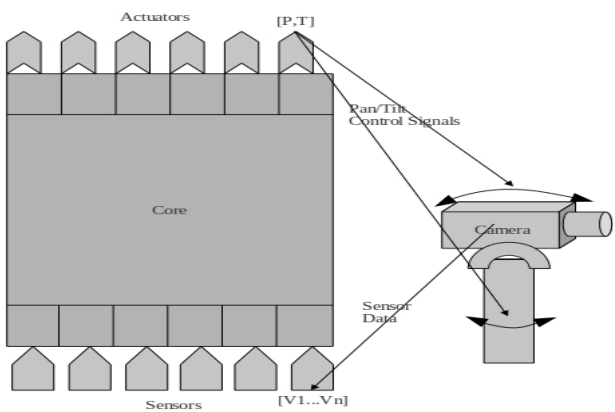

Fig. 3. DXNN gathering data from Camera, and sending pan/tilt signals to the servos.

### 3.3 SUBCORE:

SubCore is a Level 2 element. Each SubCore is a mini server/client that supervises its own NN. A SubCore accepts vectors as inputs and produces a vector as an output [Fig4]. In this paper the SubCore program primarily deals with distributing and aggregating signals to and from neurons. Each SubCore has its own Id/Address, InputList, OutputList, NeuralConnectedToList, NeuralConnectedFromList, Type, ParameterList, and a list of supervised Neuron_Ids if any. The Type variable determines the behavior of the SubCore and how/what it does with the signals it aggregates, whether it supervises a NN or whether it's a static program. The ParameterList can further modify the SubCore's functionality. When stored in a database each SubCore is represented as a tuple: {Id, InputList, OutputList, NeuralConnectedToList, NeuralConnectedFromList, Type, ParameterList, SupervisedNeuronIds, Generation}. All SubCores work in parallel and because the way they work is based on their Type, a Core can have a number of SubCores of different types working and processing and passing data to each other concurrently. In this way, a committee machine topology, or any other type of topology can be evolved. For example, one SubCore can be of type "Neural" supervising a NN, a second can have type "Substrate" and use its supervised NN to drive a neural substrate[5], finally a third SubCore can be a simple static program that does not supervise a Neural Network at all. All 3 of these SubCores might further be linked to another SubCore that collects and weighs the votes, thus producing a committee machine. By clustering the NNs into SubCore modules and distributing those modules amongst separate CPUs or machines, the hardware need only exchange the single vectors produced by the SubCores between CPUs rather than hundreds or thousands of separate smaller signals produced by each Neuron.

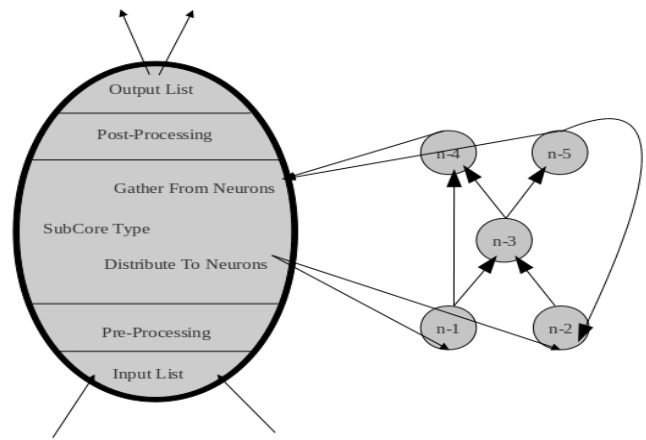

Fig. 4. SubCore accepting a vector input, distributing it to NN, gathering resulting values from the NN, and forwarding a vector output.

### 3.4 NEURON:

Neuron is a Level 1 element. A Neuron accepts vectors as inputs and outputs a resulting vector of length 1 [Fig5]. When live, the Neuron is a mini server/client program with an Id/Address, InputList, OutputList, ActivationFunction, LearningMethod, WeightList, and a ParameterList which might further augment the Neuron's functionality. When stored in a database it is represented as a tuple: {Id, InputList, OutputList, ActivationFunction, LearningMethod, WeightList, ParameterList, Generation}. Neurons can have any type of Activation Function, from sigmoid to the mexican hat function. During the initial DXNN creation process and during the topological mutation phase, a random Activation Function (AF) is chosen from a list of available AF programs represented by a list of tags. In such a list each tag is a program name that can operate on a value passed to it. Thus as soon as a new AF program is created, the name of that program can be added to the existing list as a tag, and later during a topological mutation phase be acquired by some neuron. Neurons also have a Learning Method (LM) which determines how to change the neuron's weights over time. A LM is a program which accepts 3 parameters, a current weight list, an input vector, and an AF. The output of the LM is an updated weight list and an output vector. The LM can be "none" and output the same weight list it was originally passed with an output equaling to the AF applied to the dot product of the weight list and the input vector. Alternatively, the LM can be "hebbian" and produce the output vector and a modified weight list by applying the hebbian learning algorithm and using the AF on the dot-product. Like the AF list, the LM list can also be easily expanded, letting future neurons stumble upon new LMs through mutation or acquire them when initially created. Finally, all Neurons are initially created without a bias and can acquire that bias input through mutation.

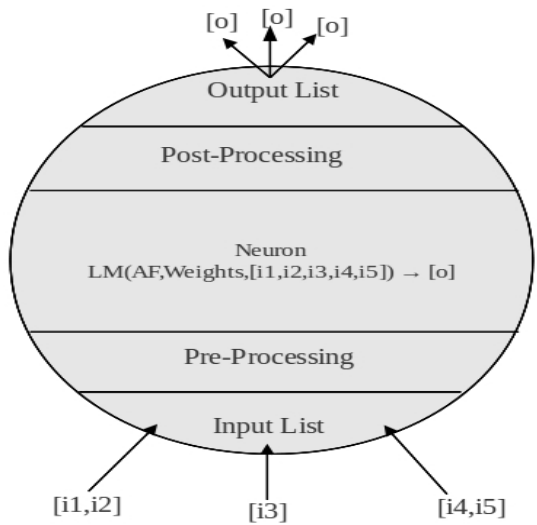

Fig. 5. A Neuron accepts input vectors, passes them with Weights and an AF to the LM, and outputs an output vector while acquiring updated weights.

Each Element (Neuron, SubCore, Core) also has a Generation variable. The Core's Generation is incremented every time it participates in the topological mutation phase, while the Generation variable of every other element is reset to that of the Core's whenever it, or one or more of its subordinates (if any), undergoes a mutation during the topological mutation phase. During the population initialization all Elements start with Generation equaling to 0. In this manner we can track how recently each Neuron and SubCore was topologically perturbed. For what further purpose the Generation and History list is used will be explained in later sections.

Putting the 3 levels together into one system [Fig1], we get the following flow of information: The Core's list of sensors produce data vectors and distribute them to the appropriate SubCores. The SubCores then process the vectors internally by passing them to the supervised neural circuits, or some other internal structures. The NNs or some other internal structures process the data and output a vector to their supervising SubCore. The SubCore, after some post-processing, passes that accumulated vector onwards to other SubCores or the Core. Finally, when the Core receives the vectors from the output layer SubCores, it passes the vectors to their appropriate actuator programs which then parse the vectors and act upon the environment. The Core then polls the sensor programs for new data vectors and the cycle of information flow repeats.

### 3.5 Representing DXNN Inside a Database:

The following encoding is used to represent DXNN within a relational database.
Population: {Population_Id, DXNN_Id_List}
DXNN: {DXNN_Id, Core_Id, ElementList}
ElementList: [ElementTuple1...ElementTupleN]
Core Element: {Id, SensorList, ActuatorList, ParameterList, SupervisedSubCoreIds, Generation, History}
SubCore Element: {Id, InputList, OutputList, NeuralConnectedToList, NeuralConnectedFromList, Type, ParameterList, SupervisedNeuronIds, Generation}
Neuron Element: {Id, InputList, OutputList, ActivationFunction, LearningMethod, WeightList, ParameterList, Generation}

Most of the elements within these tuples are lists themselves and are represented in a similar fashion. Since these are all nothing but lists of tuples, they are human readable, and very easy to store in a database and traverse through. For example, to get at any Neuron one only needs a program that asks the population for a DXNN_Id, the correlated DXNN then provides the Core_Id, the Core_Id provides a SubCore_Id, and the SubCore_Id leads to the requested Neuron_Id. In this fashion any mutation can be applied and any resulting topological perturbations due to the mutation can be followed and applied throughout the hierarchical network using these Id links.

### 4. The DXNN algorithm:

The DXNN Platform's training algorithm is divided into multiple phases. The Initialization Phase which is executed only once to create the original minimalistic population of DXNNs. The Tuning Phase where the DXNNs interact with the environment or some problem. The Selection Phase during which some DXNNs are put into the fit (valid) group and others into the unfit (invalid) group, letting only the valid DXNNs create offspring and themselves survive into the next generation. Finally followed by the Topological Mutation

Phase during which mutational operators are applied to the DXNNs, affecting topology of the SubCore Network, Neural Network, and the various non-weight parameters of the system in general. When all the phases complete, the DXNNs and their offspring (mutated versions of the valid DXNNs) are released back into the environment if the experiment is Artificial Life (AL), or applied again to the problem.

### 4.1 Initialization Phase:

During the initialization phase every element created has its Generation set to 0. Initially a population of size X is created. Each DXNN in the population starts with a minimal network, where the minimal starting topology depends on the total number of Sensors and Actuators the researcher starts the system with. If the DXNN is set to start with only 1 Sensor and 1 Actuator, then the DXNN starts with a single SubCore that contains a single layer of neurons, the number of Neurons in the single layer equals the length of the output vector destined for the Actuator. Thus if the output is a vector of length 1 like in the Double Pole Balancing (DPB) control problem, the Core contains a single SubCore which contains a single Neuron [Fig6]. If on the other hand the DXNN is initiated with N number of Sensors and and K number of actuators, the Core will contain 2 layers of fully interconnected SubCores. The first layer which contains N SubCores, and the second K SubCores. Each with a single layer of Neurons totaling the length of the output vector destined for the Actuator that the SubCore is associated with. It is preferred for the DXNNs in the population to be initialized with a single Sensor and a single Actuator, letting the DXNNs discover any other Sensors and Actuators through topological evolution.

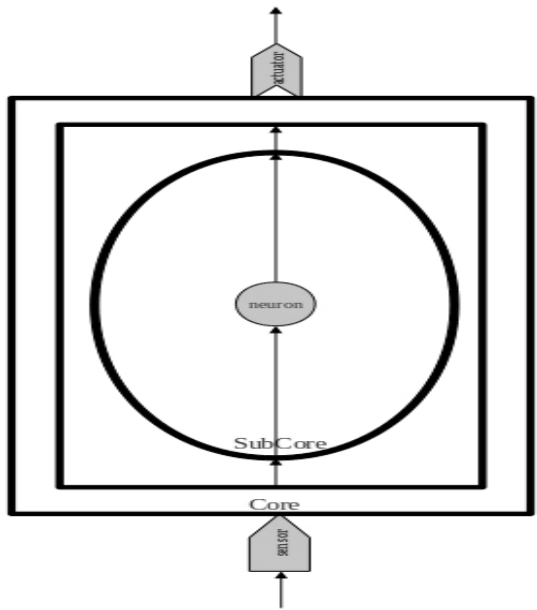

Fig. 6. A single Core containing a single SubCore containing a single Neuron.

Furthermore, the link from a SubCore to a Neuron can be of 3 types listed below. For the following examples assume that the SubCore has an Id = SubCore_Id and it receives input vectors from a Core with Id = Core_Id and a SubCore with Id = SubCore_Id2. The input vector length from Core_Id is of length 3 and looks as follows:[C1,C2,C3], whereas the input vector from SubCore_Id2 is of length 2 and looks as follows: [SC1,SC2].

Link-Types:

1) Single-type link, in which the SubCore sends the Neuron a single value from one of its input vectors. In this case the link to the Neuron within the NeuralConnectedToList is defined as a tuple: {To_Id, From_Id, single, 3}. The first element in the tuple defines an Id of the Neuron to whom the connection is made (To_Id = Neuron_Id), the second element defines an Id of the Node from whom the vector was sent to the SubCore (From_Id = Core_Id), the third element defines the Link_Type = single, and the fourth element defines the index of the value within the vector, Index = 3. Thus, this connection dictates that the SubCore is connected to a Neuron with an Id = Neuron_Id, and it will send that neuron a vector of length 1 composed of the 3rd value in the vector it receives from Core with the Id = Core_Id. Thus the Neuron will receive the following vector from the SubCore: [C3].

2) Block-type link, in which the SubCore sends the Neuron a single vector from a list of its input vectors sent to it by the Nodes in its InputList. The connection is defined as a tuple: {To_Id, From_Id, block}. Assume that To_Id = Neuron_Id and From_Id = SubCore_Id2. In this case instead of the link type being single, it is block, and so the entire vector coming from the From_Id will be routed to the Neuron. The Neuron will receive the following vector from the SubCore: [SC1,SC2].

3) All-type link, in which the SubCore sends the Neuron a single vector composed of all the vectors at its disposal concatenated together. The connection is defined as: {To_Id, all}. Here the SubCore simply gathers all the vectors in a predefined order, concatenates them together to form a single vector, and sends that vector to the Neuron. In our example, since the SubCore receives two vectors, it will send the Neuron the following single vector: [C1,C2,C3,SC1,SC2].

All this information is kept in the SubCore, the Neuron neither knows what type nor originally from whom the signal is coming. Each neuron only keeps track of the list of nodes it is connected from and the vector lengths coming from those nodes. Thus, to the Neuron all 3 of the previous link-types look exactly the same in its InputList, represented by a simple tuple {From_Id, Vector_Length}. The Vector_Length variable is of course different for each of those connections, 1 for the first, 2 for the second, and 5 for the third, but for all 3 cases the From_Id = SubCore_Id.

The different link-types add to the flexibility of the system and allow the Neurons to evolve a connection where they can concentrate on processing a single value or an entire vector coming from a Sensor, depending on the problem's need. I think this improves the general diversity of the population, allows for greater compactness to be evolved, and also improves the NN's ability to move through the fitness landscape. Since it is never known ahead of time what sensory values are needed and how they need to be processed to produce a proper output, different types of links should be allowed.

For example, a SubCore is routing to the Neurons a vector of length 100 from one of its inputs. Assume that a solution requires that a Neuron needs to concentrate on the 53rd value in the vector and pass it through a cosine activation function. To do this, the Neuron would need to evolve weights equaling to 0 for all other 99 values in the vector. This is a difficult task since zeroing each weight will take multiple attempts, and during random weight perturbations zeroing one weight might un-zero another. On the other hand evolving a single link-type to that Sensor has a 1/100 chance of being connected to the 53rd value, a much better chance. On the other hand, assume now that a solution requires for a neuron to have a connection to all of the 100 values in the vector. That is almost impossible to achieve, and would require at least 100 topological mutations if only a single link-type is used, but has a 1/3 chance of occurrence if we have block, all, and single type links at our disposal.

The SubCores themselves can also be of different types in the current system: Type = "neural" which is covered in this paper's benchmarks, and a Type = "substrate" which is not. The "neural" type SubCore is a SubCore that supervises a standard recursive Neural Network. The substrate type SubCores use their supervised NNs to drive a neural substrate, an encoding popularized by HyperNEAT. In such SubCores the input vector is routed to the substrate and the output vector comes from the substrate. The NN is polled to produce the weights for the neurons in the substrate. The substrates can differ in density and dimensionality. Examples of different substrates are shown in [Fig7].

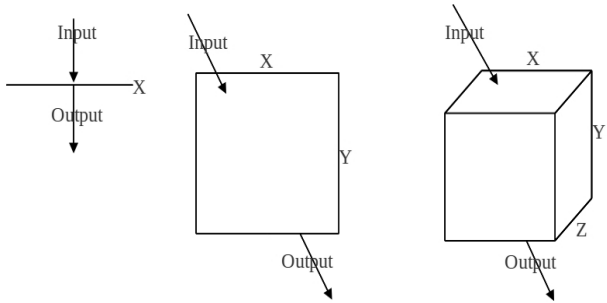

Fig. 7. Different types of substrates.

### 4.2 Tuning Phase:

The first step that must be taken is to construct/summon the Core, SubCores, and Neurons for every DXNN from the list of tuples representing these nodes within the database. The database is scanned for the {DXNN_Id, Core_Id, ElementList} tuples, each tuple has its own Id so as to identify each separate DXNN. Each ElementList which contains only the Ids of the appropriate elements is then analyzed and depending on whether the correlated tuple represents a Core, SubCore, or a Neuron, a proper independent mini server/client is summoned for each, with the parameters and links specified by the data in that tuple. The Core then composes a list of New Generation Neurons (NGN). To create an NGN list the following steps are taken:

1. Neuron Ids are sorted based on the neuron's generation, most recent(highest) to least recent(lowest).
2. Ids belonging to the 2 most recent generation are extracted, and designated CurGenIds.
3. Square root of the total number of remaining Ids is calculated, and then this number of Ids is extracted from the remaining Id list, starting from the most recent side. We designate this Id list: RecentGenIds.
4. NGN = concatenate(CurGenIds,RecentGenIds)

After NGN is composed, a variable MaxMistakes is created and set to BaseMaxMistakes + sqrt(TotWeights from NGNs) rounded to the nearest integer. The BaseMaxMistakes variable is set by the researcher. Finally, a variable by the name AttemptCounter is created and set to 1.

The reason for the creation of the NGN list is due to the weight perturbations being applied only to these new or recently modified Neurons, a method I refer to as "**Targeted Tuning**". The reason to only apply perturbations to the NGNs is because evolution in the natural world works primarily through complexification and elaboration, there is no time to re-perturb all the neurons in the network after some minor topological or other type of addition to the system. As NNs grow in size it becomes harder and harder to set all the weights and parameters of all the Neurons at the same time to such values that produces a fit individual. A system composed of a thousand neurons might have billions of parameters in it. The odds of finding proper values for them all by perturbing random weights in random Neurons throughout the entire system after some minor topological mutation, all at the same time, is slim to none. The problem only becomes more intractable as the number of Neurons grows further. By concentrating on tuning only the newly created or newly topologically/structurally augmented Neurons, and make them work with an already existing DXNN, is a much more tractable approach. Indeed in many respects it is how complexification and elaboration works in the biological NNs. In our organic brains the relatively recent evolutionary addition of the Neocortex was **not** done through some refurbishing of an older NN structure, but through a completely new addition of neural tissue covering and working with the more primordial and older parts. The Neocortex works concurrently with the older regions, contributing and when possible overwriting the signals coming from our more ancient neural structures.

During the Tuning Phase each DXNN tries to solve the problem. The DXNNs then receive fitness scores based on their performance and some fitness function for that problem. After being scored, the DXNN backs up its parameters. Afterwards, for every DXNN the Core decides on how many Neurons to mutate from its NGN list. The total number is chosen with uniform distribution to be between 1 and square root of total number of NGNs. The Core then sends these randomly chosen Neurons a request to perturb some of their weights. Each chosen Neuron when receiving such a request then perturbs its own weights. The total number of weights to be perturbed is chosen randomly by every Neuron itself. The number of weights chosen for perturbation by each neuron is a random value between 1 and square root of total number of weights in that Neuron. The perturbation value is between -(WeightLimit/2) and (WeightLimit/2), where the WeightLimit is set to Pi. By randomly selecting the total number of Neurons, the total number of weights to perturb, and then using such a wide range for the perturbation intensities, we can achieve a very wide range of parametric perturbations. Sometimes the DXNN might have only a single weight in a single Neuron perturbed slightly, while at other times it might have multiple Neurons with multiple weights perturbed to a great degree. This allows the DXNN platform to make small intensity perturbations to fine tune the parameters, but also very large intensity (number of Neurons and weights) weight perturbations which allows DXNN to jump over or out of local maximas. This would be impossible to accomplish when using only small perturbations applied to a small number of Neurons and weights. This high mutation variability method is referred to in the DXNN platform as a Random Intensity Mutation (RIM). The range of mutation intensities grows as the square root of the total number of NGNs, as it logically should since the greater the number of new or recently augmented Neurons in the DXNN, the greater the number of perturbations that needs to be applied to make a significant affect on the information processing capabilities of the system.

After all the weight perturbations have been applied within the DXNN, it attempts to solve the problem again. If the new fitness achieved by the DXNN is greater than the previous fitness, then the new weights overwrite the old backed up weights, the AttemptCounter is reset to 1, and a new set of weight perturbations is applied to the DXNN. Alternatively, if the new fitness is not greater than the previous fitness, then the old weights are restored, the AttemptCounter is incremented, and another set of weight perturbations is applied to the individual.

When the DXNN's AttemptCounter reaches the value of MaxAttempts, implying that a MaxAttempts number of RIMs have been applied and all failed to produce a higher fitness solution in sequence, the DXNN with its final best fitness and the correlated weights is backed up to the database through the conversion back to a list of tuples followed by a shut down of the DXNN itself. Utilizing the AttemptCounter and MaxAttempts strategy allows us, to some degree at least, test each topology and thus let each DXNN after the tuning phase to represent roughly the best fitness that its topology can achieve. In this way there is no need to forcefully and artificially speciate and protect the various topologies, because after the tunning phase completes, each DXNN represents roughly the highest potential that its topology can reach in a reasonable amount of time. This allows us to judge each DXNN purely on its fitness. If one increases the BaseMaxAttempt value, each DXNN will have more testing done on it with regard to weight perturbations, thus testing the particular topology more thoroughly before giving it the final fitness score. The MaxAttempt variable grows in proportion to the number of total sum of NGN weights that should be tunned, since the greater the number of new weights that need to be tuned, the more attempts it would take to properly test the various permutations. To maintain the "reasonable amount of time" clause, the MaxAttempt value is hard limited to 100.

### 4.3 SELECTION PHASE:

There are many TWEANNs that implement speciation during selection. Speciation is used to promote diversity and protect unfit individuals who in current generation do not posses enough fitness to get a chance of producing offspring or mutating and achieving better results in the future. The developer of NEAT states that new ideas need time to develop and speciation protects such innovations. Though I agree with the sentiment of giving ideas time to flourish, I consider speciation and fitness sharing or any other artificial protection of unfit individuals to be an incorrect approach, and thus refer to such speciation and other similar methods as "Forced Speciation". DXNN platform does not implement forced speciation, instead it properly tests its individuals during the Tuning Phase and utilizes natural selection during the Selection Phase that also takes into account the very complexity of the DXNNs. As in the natural world, smaller and simpler organisms require less energy and material to create offspring. As an example, for the same amount of material and energy that is required for a human to produce and raise an offspring, millions of ants can reproduce and raise offspring. When calculating who survives and how many offspring to allocate to each survivor, the DXNN platform takes complexity into account instead of blindly and artificially defending the unfit and insufficiently tested Neural Networks. Speciation and niching should be done not forcefully from the outside by the researcher, but by the artificial organisms themselves within the artificial environments they inhabit. If the organisms can indeed find their niches, thy will acquire more fitness points and secure their survival that way.

Due to the Tuning Phase, by the time the Selection Phase is reached, each individual presents its topology in roughly the best light it can reach within reasonable time. This is due to the consistent application of RIM to each DXNN, and only after a substantial number of continues failures to improve is the individual considered to be somewhere at the limits of its potential. Furthermore, when individuals are artificially protected within the population, more and more Neurons are added to the NNs unnecessarily producing the dreaded topological bloat. Topological bloating dramatically and catastrophically hinders any further improvements due to a greater number of Neurons that need to have their parameters set **concurrently** to just the right values to get the whole system functional. An example of such topological bloating was demonstrated in the robot arm control experiment using NEAT and EANT2 [7]. In that experiment NEAT continued to fail due to significant neural bloating, even though in smaller tests [3] it seemed that NEAT's speciation only improves the compactness of the solutions. Once the NN bloats past a certain size, it simply can not find a solution due to the high number of Neurons that need to have their parameters set up concurrently to a proper value. At the same time these TWEANN algorithms allow for only a small number of perturbations to be applied at any one instant.

Thus, once a NN passes some topological bloating point, it looses any chance of ever succeeding to solve the problem. In DXNN, through the use of Targeted Tuning and RIMs applied during the Tuning and Topological Mutation phases, we can successfully avoid bloating. Indeed, as will be demonstrated during the experiments in later sections, the DXNN platform consistently produces highly compact NN solutions.

### 4.4 The "Competition" Selection algorithm:

When all NNs have been given their fitness rating, we must use some method to choose from the best of those NNs. DXNN platform uses a selection algorithm I call "Competition" which tries to take into account not just the fitness of the solution, but also it's complexity. The steps of the "Competition" algorithm are as follows:

1. Calculate the average energy cost of the Neuron using the following steps:

   TotEnergy = DXNN1_Fitness + DXNN2_Fitness(2)...,

   TotNeurons = DXNN1_TotNeurons + DXNN2_TotNeurons...

   AverageEnergyCost = TotEnergy/TotNeurons

2. Sort the DXNNs in the population based on fitness. If 2 ore more DXNNs have the same fitness, they are then sorted further based on size, more compact solutions are considered of higher fitness than less compact solutions.

3. Remove the bottom 50% of the population.

4. Calculate the number of alloted offspring for each DXNN:

   AllotedNeurons = (Fitness/AverageEnergyCost),

   AllotedOffsprings(i) = round(AllotedNeurons(i)/DXNN(i)_TotNeurons)

5. Calculate total number of offspring currently being produced for the next generation:

   TotalNewOffsprings = AllotedOffsprings(1)+...AllotedOffsprings(n).

6. Calculate PopulationNormalizer, to keep the population within a certain limit:

   PopulationNormalizer = TotalNewOffsprings/PopulationLimit

7. Calculate the normalized number of offspring alloted to each DXNN:

   NormalizedAllotedOffsprings(i) = round(AllotedOffsprings(i)/PopulationNormalizer(i)).

8. If NormalizedAllotedOffsprings (NAO) == 1, then the DXNN is allowed to survive to the next generation without offspring, if NAO > 1, then the DXNN is allowed to produce (NAO -1) number of mutated copies of itself, if NAO = 0 the DXNN is removed from the population and deleted.

9. The Topological Mutation Phase is initiated, and the mutator program then passes through the database creating the appropriate (NAO) number of mutated clones of the surviving individuals.

From this algorithm it can be noted that it becomes very difficult for bloated NNs to survive when smaller systems produce better or similar results. Yet when a large NN produces significantly better results justifying its complexity, it can begin to compete and even push out the smaller NNs. This selection algorithm takes into account that a NN composed of 2 Neurons is doubling the size of a 1 Neuron NN, and thus should bring with it sizable fitness gains if it wants to produce just as many offspring. On the other hand, a NN of size 101 is only slightly larger than a NN of size 100, and thus should pay only slightly more per offspring.

### 4.5 Topological Mutation Phase:

An offspring of a DXNN is produced by first creating a clone of the parent DXNN, giving it a new unique Id, and then applying Mutation Operators to it. The Mutation Operators (MOs) that operate on the individual's topology are randomly chosen from the following list:

1. "Add Neuron" to one of the SubCores and link it randomly to and from a randomly chosen Neuron within the SubCore, or to the SubCore itself.

2. "Add Link" (can be recurrent) to or from a Neuron within the same supervising SubCore, or the SubCore itself.

2. "Splice Neuron" such that that two random Neurons which are connected to each other are disconnected and then reconnected through a newly created Neuron.

3. "Change Activation Function" of a random Neuron.

4. "Change Learning Method" of a random Neuron.

5. "Add Bias" connection (all neurons are initially created without bias).

6. If the ratio of Neurons to SubCores exceeds some value K, then the following mutation operators become available:

   1. "Add SubCore", where a new SubCore is added under Core and randomly linked to and from SubCores from a list of SubCores under the same Core, or the Core itself.

   2. "Add SubCore Link", where the link is made to or from a randomly chosen SubCore under the same Core, or the Core itself.

   3. "Splice SubCore", which chooses a random SubCore under the Core, disconnects it from another node and then reconnects it through a newly created SubCore which accommodates the appropriate Input and Output vector lengths.

The "Add SubCore" and "Add SubCore Link" can both create links (through the Core) from/to the Sensor and Actuator programs not previously used by the DXNN. In this manner the DXNN can expand its senses and control over new actuators and body parts. This feature becomes especially important when the DXNN platform is applied to the Artificial Life and Robotics experiments. The different sensors can also simply represent various features of a problem, and in this manner the DXNN platform naturally incorporates feature selection capabilities.

The total number of Mutation Operators applied to each offspring of the DXNN is a value randomly chosen between 1 and square root of the total number of Neurons in the parent DXNN. In this way once again a type of random intensity mutation (RIM) approach is utilized. Some mutant clones will only slightly differ from their DXNN parent, while others might have a very large number of Mutation Operators applied to them and thus differ drastically. This gives the offspring a chance to jump over large fitness valleys that would otherwise prove impassible if a constant number of mutational operators were to have been applied every time. As the complexity and size of each DXNN increases, each new topological mutation plays a smaller and smaller part in changing its behavior, thus a larger and larger number of mutations needs to be applied to produce significant differences to the processing capabilities of that individual. When the size of the NN is a single neuron, adding another one has a large significance. When the original size is a million neurons, adding the same single neuron to the network might not produce the same amount of change in computational capabilities of that system. Increasing the number of MOs applied when the size and complexity of the parent DXNN increases, allows us to make the mutation intensity significant enough to allow the mutant offspring to continue to produce innovations in its behavior when compared to the parent. At the same time, some offspring will only acquire a few MOs and differ topologically only slightly and thus have a chance to tune and explore the local areas on the topological fitness landscape.

Because the sensors and actuators are represented by simple lists of tags/names of existing sensor and actuator programs, the DXNN platform allows for the individuals within the population to expand their affecting and sensing capabilities through the connections formed during SubCore level mutations. Such abilities integrated naturally into the NN lets individuals gather new abilities and control over functions as they evolve. For example, originally a population of very simple individuals with only distance sensors is created. At some point when enough fitness is achieved based on some criteria, and the DXNN itself is composed of enough Neurons, the "Add SubCore" and "Add SubCore Link" mutational operators become available. When the "Add SubCore" or "Add SubCore Link" operator is randomly applied to one of the offspring of the DXNN, the SubCore has a chance of randomly linking to or from a new Sensor or Actuator. In this manner the offspring can acquire sonar or other types of sensors present in the sensor list, or acquire control of a new body part and further expand its own morphology. These types of expansions and experiments can be undertaken in the artificial life/robotics simulation environments like the Player/Stage/Gazebo Project[9]. Player/Stage/Gazebo in particular has a list of existing sensor and actuator types, making such experiments accessible at a very low cost.

Finally, once all the offspring are generated, they and their parents once more enter the tuning phase to continue with the evolutionary cycle.

### 5. SIMPLE EXPERIMENTS:

Three simple experiments will be discussed in this section. The first experiment will test whether Modular DXNN platform can evolve the topology needed to solve the XOR problem when started with a single Neuron without a bias. The second and third experiment will be that of the double pole balancing with and without velocities as specified in [4]. The results produced by Modular DXNN platform will then be compared with other state of the art TWEANNs. In each of the following experiments DXNN platform performs 100 runs with a population size limited to 10. Though DXNN benefits from using large populations, it can manage with very small populations due to the tuning phase. The parameter BaseMaxMistakes was set to 10 in the Double Pole Balancing (DPB) with velocity information experiment, and 20 in the DPB without velocity information experiments. To make the system comparable to other TWEANNs, only the hyperbolic tangent activation function was used, with the Learning Method parameter restricted to: "none".

### 5.1 XOR SIMULATION:

The minimal requirement for a TWEANN is the ability to solve the XOR benchmark starting with a single Neuron. To learn to mimic XOR it is necessary for the NN to evolve at least a single hidden Neuron, thus demonstrating its ability to evolve the necessary topology.

The DXNN platform started with single Core containing a single SubCore containing a single Neuron topologies without bias. During 100 simulations the platform was able to find the solution 100% of the time, with the system composed of 1 SubCore containing 2-3 Neurons. After having demonstrated that it could evolve rudimentary topologies, the Modular DXNN Platform was applied to the double pole balancing problems.

### 5.2. DOUBLE POLE BALANCING EXPERIMENTAL SETUP

The simulation is created using a realistic physical model incorporating friction and fourth order Runge-Kutta integration. A step size of 0.01s was used, with DXNN producing Force values at 0.02s time steps.

The state variables for the problem were as follows:
1. Cart Position
2. Cart Velocity
3. Pole1 Position
4. Pole1 Velocity
5. Pole2 Position
6. Pole2 Velocity

At every time step the DXNN receives scaled state variables from the simulation and outputs a force vector [N], where N is force. N is further scaled to be within the -10 and 10 range.

To pass the test, DXNN must balance 2 poles of different sizes (1m and 0.1m) for 100k time-steps (30 minutes of simulated time). The two poles have initial positions of 4 degrees for the long pole and 0 degrees for the short pole. Both poles must be kept within 36 degrees of the vertical. Furthermore, the cart starts at the center of a 4.8 meter track at X = 0, and must remain within -2.4 and 2.4 meters of that center throughout the experiment. Finally, the Force produced by the DXNN is set to be no less than (1/256)*10N as in [4]. The general setup of the experiment is graphically demonstrated in [Fig8].

A single experiment will run until either a maximum number of evaluations is reached, which shall be set to 50000, or until the problem is solved. The average number of evaluations during the 100 experiments will be compared to the average number of evaluations taken by other TWEANN systems. An evaluation is counted every time DXNN is given a fitness, in other words, each perturbation of weights in the Tunning Phase and subsequent application to the domain counts as an evaluation.

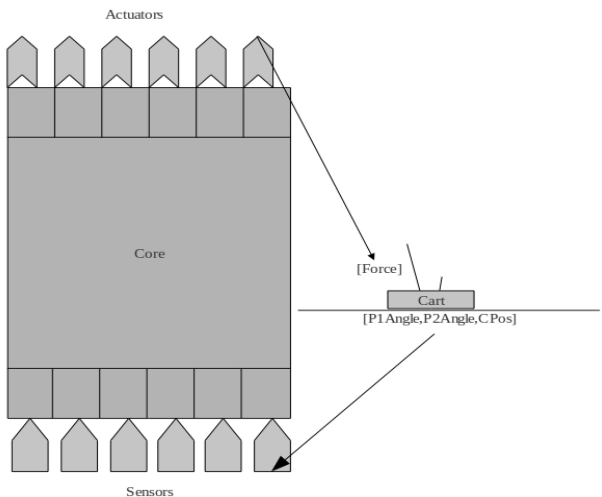

Fig. 8. DXNN gathers data from the double pole simulation and outputs a force value that is then scaled to be within -10N and 10N.

### 5.3. DOUBLE POLE BALANCING WITH VELOCITIES:

Both, the Pole Balancing Simulation and the fitness function used are made to the specifications of [4]. The data for [Table1] is taken from [Table 3] in [4], where the average number of evaluations was calculated from 50 runs.

Table1

| Method | Evaluations |
|--------|-------------|
| RWG    | 474329      |
| EP     | 307200      |
| CNE    | 22100       |
| SANE   | 12600       |
| Q-MLP  | 10582       |
| NEAT   | 3600        |
| ESP    | 3800        |
| CoSyNE | 954         |
| CMA-EX | 895         |
| DXNN   | 703         |

All differences are statistically significant ($p < 0.001$).

As can be noted by the results, DXNN platform outperforms all other systems, even those that did not have to evolve a topology. The DXNN sizes ranged from 1-2 Neurons, highly compact.

### 5.4 DOUBLE POLE BALANCING WITHOUT VELOCITIES:

The DPB without velocities is a significantly more complex problem, requiring a recurrent NN to be evolved.
Both, the Pole Balancing Simulation and the fitness function used are made to the specifications of [4]. The data for [Table2] is taken from [Table 4] in [4] where the average number of evaluations was calculated from 50 runs.

Table2

| Method | Without-Damping | With-Damping |
|--------|-----------------|--------------|
| RWG    | 415209          | 1232296      |
| SANE   | 262700          | 451612       |
| CNE    | 76906           | 87623        |
| ESP    | 7374            | 26342        |
| NEAT   |                 | 6929         |
| CMA-ES | 3521            | 6061         |
| CoSyNE | 1249            | 3416         |
| DXNN   | 2341            | 2394         |

All differences are statistically significant ($p < 0.001$).

Using the undamped fitness function the DXNN platform produced highly competitive solutions with DXNN sizes ranging between 2 - 3 Neurons. DXNN Platform lost in its evaluation count only to CoSyNE, which did not have to evolve a topology. When damped fitness function was implemented, the DXNN sizes stayed between 2 - 3 neurons. These results demonstrate that DXNN platform outperforms other state of the art TWEANN systems, and in some cases even the topologically static methods, all the while consistently producing highly compact solutions.

Based on these results the DXNN Platform is shown to produce results much faster than other topology evolving algorithms. The topological compactness can be further increased by simply increasing the Base_MaxAttempts parameter. By setting this parameter to 100, DXNN Platform produces NN solutions composed of only 2 neurons almost exclusively.

## 6. Discussion

Due to the pole balancing problems only requiring a single sensor and a single actuator, Modular DXNN effectively acted as the standard DXNN [14], with expectedly similar results. Though the new modular architecture and the platform now allows for easy construction of explicitly modular systems, it is left up to the future multi sensor and multi actuator tests to determine the benefits of dynamic explicit modularity, and in what situations such modularity should be utilized. Using the new explicitly modular architecture, evolving multiple monolithic solutions, and then combining them as parallel SubCores under a single Core, can already be done for the purpose of hand crafting efficient committee machines, without any further modifications to the architecture or the system in general.

An example of possible advantages of explicitly modular NNs is as follows:
1. Simpler distribution of neural circuits among distributed hardware.
2. Ability to efficiently design committee machines, by first training monolithic DXNNs, and then arranging the monolithic NNs as a set of parallel SubCores within a Core, with the output layer SubCore acting as a static program which takes in votes and outputs a weighted average result. This can also be done dynamically through various topological and mutation constraints.
3. Sensor and Actuator isolation. There should be times when it is pertinent to have a NN dedicated to analyzing data coming from a single sensor, before that signal is sent onwards to be processed with other sensory input by other NN structures. For such problems, explicit modularity would provide an advantage.

Possible disadvantages to be tested for in the future:

1. The "Translation Problem"; each module outputs a vector, and accepts vectors from other modules. Whenever a module undergoes mutation, the vector it outputs becomes very different, and other modules need to relearn how to interpret these new vectors, thus they in turn too have to evolve... This might lead to inefficient evolution.
2. The question of when to allow the system to create new modules/subcores is an important question. If the system is allowed to create new subcores during every topological mutation phase, NNs which are composed of many SubCores each containing a single Neuron might evolve, instead of a standard monolithic NN. Thus some a set of constraints most likely will be necessary to prevent inefficient evolution.

## 7. Summary and Conclusion

In this paper I presented Modular DXNN, a novel Topology and Weight Evolving Artificial Neural Network platform that utilizes explicit modularity and hierarchy. DXNN uses a database friendly, tuple based, human readable, and non analog encoded NN representation. DXNN demonstrated superior performance in double pole balancing experiment with and without velocity information. The extension into modular and hierarchical topology opens a number of interesting possibilities to the platform. An efficient utilization of explicit modularity, and the conditions under which it should be utilized, will be the target of future experiments and simulations. Though in the end it might be found that monolithic systems that naturally evolve modularity are superior to the explicitly modular systems, which is my suspicion; this "Explicit Modularity" extension will surely participate in the DXNN's ability to illuminate the importance of modularity, and the types of modular approaches that would yield the best results.